\documentclass[sigconf]{acmart}
\usepackage{caption}
\usepackage{graphicx}
\usepackage{subcaption}
\usepackage{hyperref}
\usepackage{amsfonts}
\usepackage{url}
\usepackage{multirow}
\usepackage{makecell}
\usepackage{color}
\usepackage{xspace}
\usepackage{amsmath}
\usepackage{adjustbox}
\usepackage{booktabs}
\usepackage{algorithm}
\usepackage{algpseudocode}
\usepackage{soul}
\usepackage{float}
\usepackage{color}
\usepackage{enumitem}

\newcommand{\model}{\textbf{LLM4DyG} }
\newcommand{\modelnosp}{\textbf{LLM4DyG}}
\newcommand{\modelv}{\textbf{DST2} }
\newcommand{\modelvnosp}{\textbf{DST2}}

\newcommand{\ms}[2]{{#1}\scriptsize{$\pm$#2}}
\newcommand{\msone}[2]{\bf {#1}\scriptsize{$\pm$#2}}
\newcommand{\mstwo}[2]{\underline{{#1}\scriptsize{$\pm$#2}}}

\newcommand{\eg}{\textit{ e.g.}}
\newcommand{\etc}{\textit{ etc}}

\newcommand{\todo}[1]{%
    
}

\usepackage{tabularx}
\usepackage[normalem]{ulem}
\useunder{\uline}{\ul}{}
\AtBeginDocument{%
  \providecommand\BibTeX{{%
    \normalfont B\kern-0.5em{\scshape i\kern-0.25em b}\kern-0.8em\TeX}}}

\copyrightyear{2024}
\acmYear{2024}
\setcopyright{rightsretained}
\acmConference[KDD '24]{Proceedings of the 30th ACM SIGKDD Conference on Knowledge Discovery and Data Mining}{August 25--29, 2024}{Barcelona, Spain}
\acmBooktitle{Proceedings of the 30th ACM SIGKDD Conference on Knowledge Discovery and Data Mining (KDD '24), August 25--29, 2024, Barcelona, Spain}\acmDOI{10.1145/3637528.3671709}
\acmISBN{979-8-4007-0490-1/24/08}

\makeatletter
\gdef\@copyrightpermission{
  \begin{minipage}{0.3\columnwidth}
  \href{https://creativecommons.org/licenses/by/4.0/}{\includegraphics[width=0.90\textwidth]{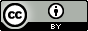}}
  \end{minipage}\hfill
  \begin{minipage}{0.7\columnwidth}
  \href{https://creativecommons.org/licenses/by/4.0/}{This work is licensed under a Creative Commons Attribution International 4.0 License.}
  \end{minipage}
  \vspace{5pt}
}
\makeatother

\begin{document}

\title{LLM4DyG: Can Large Language Models Solve Spatial-Temporal Problems on Dynamic Graphs?}
\renewcommand{\shorttitle}{LLM4DyG: Can Large Language Models Solve Spatial-Temporal Problems on Dynamic Graphs?}


\author{Zeyang Zhang}
\orcid{0000-0003-1329-1313}
\affiliation{%
  \institution{DCST, Tsinghua University}
  \city{Beijing}
  \country{China}
}
\email{zy-zhang20@mails.tsinghua.edu.cn}

\author{Xin Wang}\authornotemark[1]
\orcid{0000-0002-0351-2939}
\affiliation{%
  \institution{DCST, BNRist, Tsinghua University}
  \city{Beijing}
  \country{China}
}
\email{xin\_wang@tsinghua.edu.cn}

\author{Ziwei Zhang}
\orcid{0000-0003-2451-843X}
\affiliation{%
  \institution{DCST, Tsinghua University}
  \city{Beijing}
  \country{China}
}
\email{zw-zhang16@tsinghua.org.cn}

\author{Haoyang Li}
\orcid{0000-0003-3544-5563}
\affiliation{%
  \institution{DCST, Tsinghua University}
  \city{Beijing}
  \country{China}
}
\email{lihy218@gmail.com}

\author{Yijian Qin}
\orcid{0000-0002-0419-5226}
\affiliation{%
  \institution{DCST, Tsinghua University}
  \city{Beijing}
  \country{China}
}
\email{qinyj19@mails.tsinghua.edu.cn}

\author{Wenwu Zhu}\authornote{Corresponding Authors.}
\orcid{0000-0003-2236-9290}
\affiliation{%
  \institution{DCST, BNRist, Tsinghua University}
  \city{Beijing}
  \country{China}
}
\email{wwzhu@tsinghua.edu.cn}

\renewcommand{\shortauthors}{Zeyang Zhang et al.}

\begin{abstract}
  In an era marked by the increasing adoption of Large Language Models (LLMs) for various tasks, there is a growing focus on exploring LLMs' capabilities in handling web data, particularly graph data. Dynamic graphs, which capture temporal network evolution patterns, are ubiquitous in real-world web data. Evaluating LLMs' competence in understanding spatial-temporal information on dynamic graphs is essential for their adoption in web applications, which remains unexplored in the literature. 
In this paper, we bridge the gap via proposing to evaluate LLMs' spatial-temporal understanding abilities on dynamic graphs, to the best of our knowledge, for the first time. Specifically, we propose the \model benchmark, which includes nine specially designed tasks 
considering the capability evaluation of LLMs from both temporal and spatial dimensions. 
Then, we conduct extensive experiments to analyze the impacts of different data generators, data statistics, prompting techniques, and LLMs on the model performance. 
Finally, we propose Disentangled Spatial-Temporal Thoughts (\modelvnosp) for LLMs on dynamic graphs to enhance LLMs' spatial-temporal understanding abilities.
Our main observations are: 1) LLMs have preliminary spatial-temporal understanding abilities on dynamic graphs, 2) Dynamic graph tasks show increasing difficulties for LLMs as the graph size and density increase, while not sensitive to the time span and data generation mechanism, 3) the proposed \modelv prompting method can help to improve LLMs' spatial-temporal understanding abilities on dynamic graphs for most tasks.
The data and codes are publicly available at \href{https://github.com/wondergo2017/LLM4DyG}{Github}.
\end{abstract}

\ccsdesc[300]{Information systems~Data mining}
\ccsdesc[100]{Computing methodologies~Natural language processing}
\ccsdesc[300]{Computing methodologies~Knowledge representation and reasoning}


\keywords{Dynamic Graph; Large Language Model; Spatial-Temporal; Benchmark; Evaluation; Disentanglement}



\maketitle
\section{Introduction}
In an era marked by the increasing adoption of Large Language Models (LLMs) for various tasks beyond natural language processing, such as image recognition~\cite{alayrac2022flamingo}, healthcare diagnostics~\cite{thirunavukarasu2023large}, and autonomous agents~\cite{wang2023survey}, there has been a growing body of research dedicated to exploring LLMs' abilities to tackle the vast troves of web data. One area of particular interest is the handling of graph data, which ubiquitously exists on the Internet. The World Wide Web itself can be seen as a colossal interconnected graph of webpages, hyperlinks, and content. For example, social media platforms like Facebook, Twitter, and Instagram generate dynamic social graphs reflecting user interactions and connections.

To leverage the in-context learning and commonsense knowledge of LLMs, several pioneer works have been dedicated to adopting LLMs on static graphs.
For instance, ~\citet{wang2023can} and ~\citet{guo2023gpt4graph} propose benchmarks to evaluate LLMs' proficiency in comprehending and reasoning about graph structures, with tasks like graph connectivity, topological sort,\etc, demonstrating the LLMs' abilities of in-context learning and reasoning to solve static graph problems. ~\citet{ye2023natural} and ~\citet{chen2023exploring} propose to fine-tune the LLMs to solve graph tasks in natural language, showing the strong potential of LLMs to leverage text information, generate human-readable explanations, and integrate commonsense knowledge to enhance the reasoning over structures.

Dynamic graphs, in comparison with static graphs, possess a wealth of temporal evolution information, which is more prevalent on the internet. For instance, on platforms such as Twitter, users engage in continuous interactions with each other, and on Wikipedia, knowledge graphs are kept updated over time.  
On the one hand, with the additional temporal dimension, it is possible for LLMs to interpret the ever-changing relationships and information updates on dynamic graphs, which are ignored in static graphs. On the other hand, there exist additional research challenges for capturing the graph dynamics, and evaluating LLMs' proficiency in comprehending spatial-temporal information is critical for the applications of LLMs on dynamic graphs. Such investigations hold the potential to shed light on broader web applications such as sequential recommendation, trend prediction, fraud detection,\etc.

To this end, in this paper, we propose to explore the following research question: 
\\

\textit{Can large language models understand and handle the spatial-temporal information on dynamic graphs in natural language?} 
\\

However, this problem remains unexplored in literature, and is non-trivial with the following challenges: 

\begin{itemize}[leftmargin=0.5cm]
    \item How to design dynamic graph tasks to assess the capabilities of LLMs to understand temporal and structural information both separately and simultaneously.
    \item How to investigate the impacts of spatial and temporal dimensions, where they have complex and mixed interactions on dynamic graphs.
    \item How to design the prompts for dynamic graphs and tasks, where spatial-temporal information should be taken into consideration in natural language.
\end{itemize}

\begin{figure*}
    \centering
    \includegraphics[width = 0.981\textwidth]{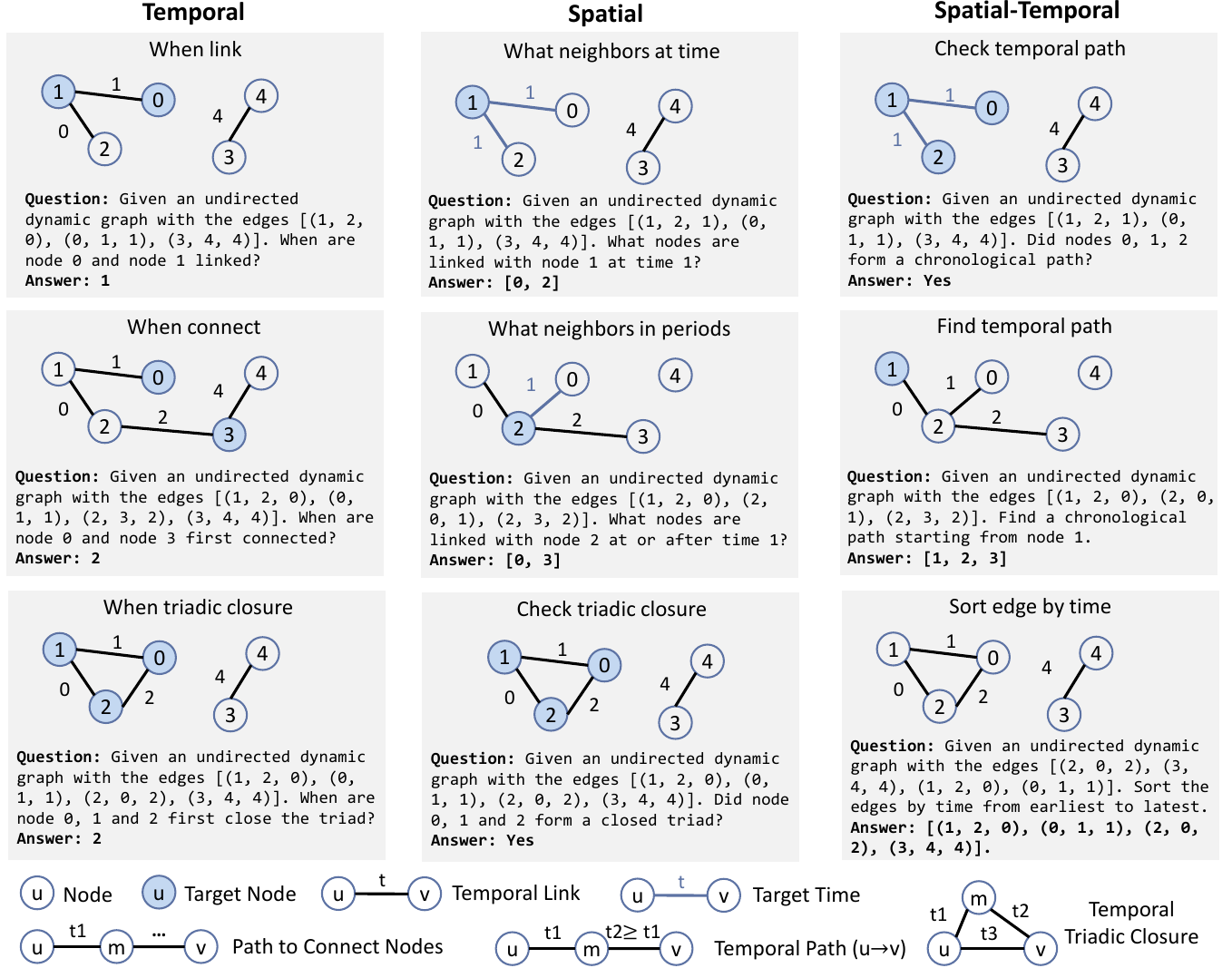}
    \caption{An overview of the tasks in the \model Benchmark. The tasks are designed to consider both temporal and spatial dimensions, and question LLMs in natural language {\it when, what} or {\it whether} the spatial-temporal patterns take place. The spatial-temporal patterns range from temporal links, and chronological paths to dynamic triadic closure. The tasks are classified based on the targets of the queries. An example prompt and graph illustration are provided for each task.}
    \label{fig:task}
\end{figure*}

To address these issues, we further propose \modelnosp, a comprehensive benchmark for evaluating the spatial-temporal understanding abilities of LLMs on dynamic graphs. Specifically, we design nine specially designed tasks (illustrated in Figure~\ref{fig:task}) that consider the capability evaluation from both temporal and spatial dimensions, and question LLMs {\it when, what} or {\it whether} the spatial-temporal patterns, ranging from temporal links, and chronological paths to dynamic triadic closure, take place. To obtain a deeper analysis of the impacts of spatial and temporal dimensions for LLMs on dynamic graphs, we make comparisons on these tasks with three data generators (including Erdős-Rényi model~\cite{erdHos1960evolution}, stochastic block model~\cite{holland1983stochastic}, and forest fire model~\cite{leskovec2007graph}), various data statistics (including time span, graph size and density), four general prompting techniques (including zero/one-shot prompting, zero/one-shot chain-of-thoughts prompting~\cite{wei2022chain}), and five LLMs (including closed-source GPT-3.5 and open-source LLMs Vicuna-7B, Vicuna-13B~\cite{vicuna2023}, Llama-2-13B~\cite{touvron2023llama}, and CodeLlama-2-13B~\cite{rozière2023code}). Inspired by the observations and dynamic graph learning literature, we further design a dynamic graph prompting technique, i.e., Disentangled Spatial-Temporal Thoughts (\modelvnosp), to encourage LLMs to process spatial and temporal information sequentially. We observe the following findings from conducting extensive experiments with \model:
\begin{enumerate}[leftmargin=0.5cm]
    \item \textbf{LLMs have preliminary spatial-temporal understanding abilities on dynamic graphs.} We find that LLMs significantly outperform the random baseline on the dynamic graph tasks, and the improvements range from +9.8\% to +73\% on average in Table~\ref{tab:main}, which shows that LLMs are able to recognize structures and time, and to perform reasoning in dynamic graph tasks.
    \item \textbf{Dynamic graph tasks exhibit increasing difficulties for LLMs as the graph size and density grow, while not sensitive to the time span and data generation mechanism.}  Specifically, the performance of GPT-3.5 in the `when link' task drops from 48\% to 27\%  when the density increases from 0.3 to 0.7, while the performance varies slightly as the time span changes for most tasks in Figure~\ref{fig:Tp}. We also find that in the `when connect' task the performance drops from 97.7\% to 17.7\% when the graph size increases from 5 to 20 in Table~\ref{tab:main}. 
    \item \textbf{Our proposed \modelv prompting technique can help LLMs to improve spatial-temporal understanding abilities.} 
    We find that the results of the existing prompting techniques vary a lot for different tasks in Table~\ref{tab:prompt}. Inspired by dynamic graph literature, our proposed \modelv encourages LLMs to first consider time before nodes, thus improving the performance for most tasks, particularly, from 33.7\% to 76.7\% in the `when link' task in Table~\ref{tab:adv_prompt}. 
\end{enumerate}

To summarize, we make the following contributions:

\begin{itemize}[leftmargin=0.5cm]
    \item We propose to evaluate LLMs' spatial-temporal understanding capabilities on dynamic graphs for the first time, to the best of our knowledge.
    \item We propose the \model benchmark to comprehensively evaluate LLMs on dynamic graphs. \model consists of nine dynamic graph tasks in natural language with considerations of both temporal and spatial dimensions, ranging from temporal links, and chronological paths to dynamic triadic closure and covering questions regarding {\it when, what} or {\it whether} for LLMs. 
    \item We conduct extensive experiments taking into account three data generators, three graph parameters, four general prompts, and five different LLMs. Based on the experiments, we provide fine-grained analyses and observations about the evaluation of LLMs on dynamic graphs. 
    \item We propose a Disentangled Spatial-Temporal Thoughts (\modelvnosp) prompting technique. Experimental results show that it can greatly improve the spatial-temporal reasoning ability of LLMs. 
\end{itemize}

\section{Related Work}
\subsection{LLMs for tasks with graph data}

Recently, there has been a surge of works about LLMs for solving tasks with graph data~\cite{zhang2023large, zhang2023graph,li2023survey,jin2023large}. \citet{he2023explanations} proposed an approach that LLMs not only execute zero-shot predictions but also generate coherent explanations for their decisions. These explanations are subsequently leveraged to enhance the features of graph nodes for node classification in text-attributed graphs. ~\citet{chen2023exploring} proposes to explore LLMs-as-Enhancers and LLMs-as-Predictors for solving graph-related tasks, where the former augment the GNN with LLMs, and the latter directly adopts LLMs to make predictions. \citet{wang2023can} introduced NLGraph, a benchmarking framework tailored for evaluating the performance of LLMs on traditional graph-related tasks. Simultaneously, ~\citet{guo2023gpt4graph} conducted a comprehensive empirical study focused on utilizing LLMs to tackle structural and semantic understanding tasks within graph-based contexts. Recent contributions in this line of research include InstructGLM \cite{ye2023natural}, a method for fine-tuning LLMs inspired by LLaMA~\citep{touvron2023llama}, designed specifically for node classification tasks. ~\citet{zhang2023toolformer} and ~\citet{jiang2023structgpt} have initiated the exploration of this frontier by interfacing LLMs with external tools and enhancing their reasoning capabilities over structured data sources such as knowledge graphs (KGs) and tables. ~\citet{yao2024exploring} explores leveraging LLMs for graph generation, which have potentials for various real-world tasks like drug discoveries. ~\citet{li2024llm} utilizes LLMs to recognize and provide the possible latent causal structures given the rich domain knowledge in the textual corpus. 
However, these works mainly focus on static graphs, ignoring the temporal nature of graphs in real-world web applications. In this paper, we propose to explore LLMs' spatial-temporal understanding on dynamic graphs, which remains unexplored in the literature.

\subsection{LLMs for other related tasks}
LLMs have been recently applied to other related tasks, including time-series forecasting, recommendation, \etc. 
~\citet{yu2023temporal} presents a novel study on harnessing LLMs' outstanding knowledge and reasoning abilities for explainable financial time series forecasting.~\citet{chang2023llm4ts}  leverages pre-trained LLMs to enhance time-series forecasting and has shown exceptional capabilities as both a robust representation learner and an effective few-shot learner. ~\citet{sun2023test} summarizes two strategies for completing time-series (TS) tasks using LLM: LLM-for-TS that designs and trains a fundamental large model for TS data and TS-for-LLM that enables the pre-trained LLM to handle TS data. ~\citet{feng2023llm4vg} and ~\cite{zhang2024large2} evaluate the temporal or sequential understanding abilities of LLMs in visual tasks. ~\citet{lyu2023llm} investigates various prompting strategies for enhancing personalized recommendation performance with large language models through input augmentation. However, these works do not consider the role of structures, and in this paper, we mainly focus on exploring the spatial-temporal understanding abilities of LLMs on dynamic graphs.

\begin{figure*}
    \centering
    \includegraphics[width = 0.99\textwidth]{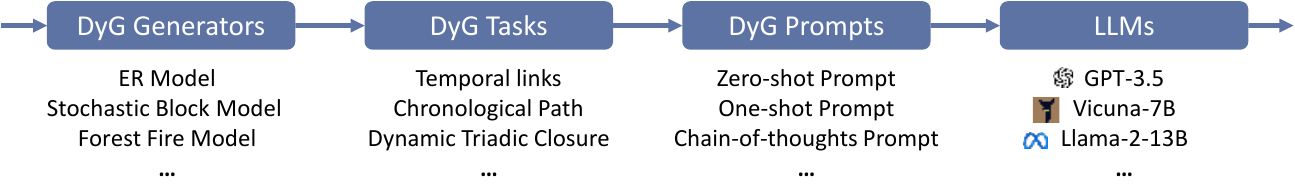}
    \caption{An overview of the pipeline in the \model Benchmark, which includes various dynamic graph generators, tasks, prompt methods, and LLMs for evaluation. }
    \label{fig:pipeline}
\end{figure*}

\subsection{Dynamic Graph Learning}
Dynamic graphs are pervasive in a multitude of real-world applications, spanning areas such as event forecasting, recommendation systems, \etc~\cite{cai2021structural,deng2020dynamic,you2019hierarchical,wang2021tedic,li2019fates,wu2020temp}. This prevalence has prompted significant research interest in the development and refinement of dynamic graph neural networks~\cite{skarding2021foundations,zhu2022learnable,chen2023easydgl,zhang2023dynamic}. These networks are designed to model intricate graph dynamics, which incorporate evolving structures and features over time. A variety of approaches have been proposed to address the challenges posed by dynamic graphs. Some research efforts have focused on employing Graph Neural Networks (GNNs) to aggregate neighborhood information for each individual snapshot of the graph. Subsequently, these methods use a sequence module to capture and model the temporal information~\cite{yang2021discrete,sun2021hyperbolic,hajiramezanali2019variational,seo2018structured,sankar2020dysat}. In contrast, other studies have proposed the use of time-encoding techniques. These methods encode the temporal links into specific time-aware embeddings, and then utilize a GNN or memory module~\cite{wang2021inductive,cong2021dynamic,xu2020inductive,rossi2020temporal} to process and handle the structural information embedded in the graph. Recently, there are some works focusing on studying dynamic graphs under distribution shifts~\cite{zhang2023ood,zhang2023outofdistribution,zhang2023spectral,zhang2022dynamic}.
However, these methods require the model to be trained  
every time they encounter a new dynamic graph task, limiting their widespread usage in real-world scenarios. In this paper, we explore the potential of LLMs on solving dynamic graph tasks with in-context learning skills and evaluate their spatial-temporal understanding abilities on dynamic graphs.

\section{The \model Benchmark}

In this section, we introduce our proposed \model benchmark to evaluate whether LLMs are capable of understanding spatial-temporal information on the dynamic graph. Specifically, we first adopt a random dynamic graph generator to generate the base dynamic graphs with controllable parameters like time span. Then, we design nine dynamic graph tasks to evaluate LLMs' abilities considering both spatial and temporal dimensions. The overall pipeline is illustrated in Figure~\ref{fig:pipeline}.
Based on this pipeline, we can control the data generation, statistics, prompting methods, and LLMs for each task to conduct fine-grained analyses. 

\subsection{Dynamic Graph Data Generators}
We first adopt a random dynamic graph data generator to control the statistics of the dynamic graph. In default, we adopt an Erdős-Rényi (ER) model to generate an undirected graph, and randomly assign a time-stamp for each edge. Denote a graph $\mathcal{G} = (\mathcal{V}, \mathcal{E})$ with the node set $\mathcal{V} = \{v_1, v_2, \dots v_N\}$ and edge set $\mathcal{E}=\{e_1, e_2, \dots, e_M \}$. We first generate the graph with the ER model $\mathcal{G} = ER(N,p)$ where $N$ is the number of nodes in the graph, and $p$ is the probability of edge occurrence between each node pair. In this way, $N$ controls the graph size, and $p$ controls the graph density. After obtaining the graph $\mathcal{G}$, we assign each edge with a random timestamp $t\sim U(\{0,1,\dots, T-1\})$, where $T$ controls the time span. For a generated dynamic graph, each edge $e = (v_i, v_j, t)$ denotes that node $v_i$ and node $v_j$ are linked at time $t$. We also include other dynamic graph generators, stochastic block (SB) model, and forest fire (FF) model. For real-world datasets, we adopt a random sampler to sample ego-graphs from the real graphs for evaluation.

\subsection{Dynamic Graph Tasks}
To evaluate LLMs' spatial-temporal understanding abilities, we design nine tasks considering both temporal and spatial dimensions. The tasks are classified based on the targets of the queries, \eg, the temporal tasks make queries about the time, the spatial tasks make queries about the nodes, while the solutions in spatial-temporal tasks are more complex and include the spatial-temporal patterns mixed together. We introduce the definition and generation of each task as follows. 

\begin{itemize}[leftmargin = 0.5cm]
    \item \textbf{Temporal Task 1: when link.} We ask when two nodes are linked in this task. In a dynamic graph $\mathcal{G} = (\mathcal{V}, \mathcal{E}) $, two nodes $u$ and $v$ are linked at time $t$ if there exists a temporal edge $(u,v,t)$ in the edge set $\mathcal{E}$. We randomly select an edge from the edge set as the query.
    \item \textbf{Temporal Task 2: when connect.} We ask when two nodes are connected in this task. In a dynamic graph $\mathcal{G} = (\mathcal{V}, \mathcal{E})$, two nodes $u$ and $v$ are connected at time $t$ if there exists a path $[(u, k_1, t), (k_1, k_2, t), \dots , (k_i, v, t)]$ from node $u$ to node $v$ at time $t$ in the edge set $\mathcal{E}$. We randomly select a pair of nodes that are connected at some time as the query.
    \item \textbf{Temporal Task 3: when triadic closure (tclosure).} We ask when the three given nodes first form a closed triad in this task. Dynamic triadic closure has been shown critical for dynamic graph analyses~\cite{zhou2018dynamic}. In a dynamic graph $\mathcal{G} = (\mathcal{V}, \mathcal{E})$, two nodes with a common neighbor are said to have a triadic closure, if they are linked since some time so that the three nodes have linked with each other to form a triad. We randomly select a closed triad as the query.
\end{itemize}

Note that while these temporal tasks focus on making queries about time, they also require the model to understand structures so that the model can recognize {\it when} some structural patterns exist, from links and paths to dynamic triads. Next, we introduce the spatial tasks that require the model to spot the specific time and discover the structures.

\begin{itemize}[leftmargin = 0.5cm]
    \item \textbf{Spatial Task 1: what neighbor at time.} In this task, we ask what nodes are linked with a given node at a given time. We randomly select a time and a node not isolated in the time-related graph snapshot to construct the query. 
    \item \textbf{Spatial Task 2: what neighbor in periods.} In this task, we ask what nodes are linked with a given node after or at a given time, but not linked before the given time. We randomly select a time and a node not isolated before the given time to construct the query. This task measures the model's abilities to understand structures within a time period, \eg, the latest links.
    \item \textbf{Spatial Task 3: check triadic closure (tclosure).} We ask whether the three given nodes form a closed triad in the dynamic graph through true/false questions. We uniformly sample from the sets of closed triads and open triads to construct the positive and negative samples respectively. We also keep a balanced number of positive and negative samples in the dataset. 
\end{itemize}

Similarly, these tasks also require the model to spot the time, from a specific time and time period to the full-time span, and then to recognize {\it what } structural patterns or {\it whether} the given structural patterns meet the requirements of the queries. Next, we introduce the spatial-temporal tasks that directly require the LLMs to process the spatial-temporal targets.

\begin{itemize}[leftmargin = 0.5cm]
    \item \textbf{Spatial-Temporal Task 1: check temporal path (tpath).} In this task, we ask whether the given three ordered nodes form a chronological path. In a dynamic graph $\mathcal{G} = (\mathcal{V}, \mathcal{E})$, a sequence of nodes $[v_1,v_2,\dots,v_n]$ construct a chronological path if the timestamps of the edges do not decrease from source node $v_1$ to target node $v_n$ in the path. We randomly select positive and negative samples from the set of chronological paths and non-chronological paths to construct a balanced dataset.
    
    \item \textbf{Spatial-Temporal Task 2: find temporal path (tpath).} In this task, we ask the model to find a chronological path starting from a given node in the dynamic graph. We randomly select a node that is a starting node at any chronological path to construct the queries. Note that any valid chronological path starting at the given node is a correct answer.
    
    \item \textbf{Spatial-Temporal Task 3: sort edge by time.} In this task, we shuffle the edges and ask the model to sort the edges by time from earliest to latest. In the cases where some edges have the same timestamp, the orders within these edges do not matter for the correct answers.
\end{itemize}

These tasks require the model to understand the spatial-temporal information at the local or global scale. The targets of the queries include both temporal and spatial information on the dynamic graph. The example prompts and illustrations are shown in Fig.~\ref{fig:task}.

\section{Experiments}
In this section, we conduct experiments to evaluate LLMs' spatial-temporal understanding abilities on dynamic graphs. We conduct fine-grained analyses with various settings from different aspects, including data, prompting methods, models, \etc. 

\begin{table}[]
\caption{An example of prompt construction for the `when connect' task.}
\label{tab:example}
\begin{tabularx}{0.48\textwidth}{lX}
\toprule
Prompt             & Example                                                                                                                                                                                                                                                                                                        \\ \midrule
DyG Instruction    & In an undirected dynamic graph, (u, v, t) means that node u and node v are linked with an undirected edge at time t.                                                                                                                                                                                           \\ \midrule
Task Instruction   & Your task is to answer when two nodes are first connected in the dynamic graph. Two nodes are connected if there exists a path between them.                                                                                                                                                                   \\ \midrule
Answer Instruction & Give the answer as an integer number at the last of your response after 'Answer:'                                                                                                                                                                                                                              \\ \midrule
Exemplar           & Here is an example: Question: Given an undirected dynamic graph with the edges [(0, 1, 0), (1, 2, 1), (0, 2, 2)]. When are node 0 and node 2 first connected? Answer:1                                                                                                                                         \\ \midrule
Question           & Question: Given an undirected dynamic graph with the edges [(0, 9, 0), (1, 9, 0), (2, 5, 0), (1, 2, 1), (2, 6, 1), (3, 7, 1), (4, 5, 2), (4, 7, 2), (7, 8, 2), (0, 1, 3), (1, 6, 3), (5, 6, 3), (0, 4, 4), (3, 4, 4), (3, 6, 4), (4, 6, 4), (4, 9, 4), (6, 7, 4)]. When are node 2 and node 1 first connected? \\ \midrule
Answer             & Answer:1      \\ \bottomrule
\end{tabularx}
\vspace{-10pt}
\end{table}

\subsection{Setups}
\paragraph{Random baseline} To verify whether the model can understand dynamic graphs instead of outputting random answers, we adopt a random baseline that uniformly selects one of the possible solutions as the answer. The accuracies of the random baseline for the tasks can be calculated by the ratio of the number of correct solutions over the number of possible solutions, which are specifically provided as follows. For the tasks `when connect' and `when tclosure', the baseline accuracy is $\frac1T$. For the task `when link', the baseline accuracy is $\frac{1}{\sum_i^T C(T,i)}$, where $C(T,i)$ is the combination number. For the tasks `neighbor at time' and`neighbor in periods', the baseline accuracy is $\frac{1}{\sum_i^N C(N,i)}$. For the tasks `check tclosure' and`check tpath', the baseline accuracies are 1/2, as the answer is either no or yes. For the tasks `find tpath' and `check tpath', the baseline accuracies are calculated by enumerating possible solutions and correct solutions for each instance. 

\paragraph{Prompting methods} To investigate how different prompting techniques affect the model's abilities, we compare various prompting methods, including zero-shot prompting, few-shot prompting~\cite{GPT3}, chain-of-thought prompting (COT)~\cite{wei2022chain} and few-shot prompting with COT. We adopt one example for few-shot prompting, and use one-shot prompting as the default prompting approach. For each problem instance, the prompt is constructed by sequentially concatenating dynamic graph instruction, task instruction, answer instruction, exemplar prompts, and question prompts. An example of the prompt construction is shown in Table~\ref{tab:example}.  

\paragraph{Models} We use GPT-3.5-turbo-instruct as the default LLM, and we also include other LLMs like Vicuna-7B, Vicuna-13B, Llama-2-13B and CodeLlama-2-13B. For all models, we set temperature $\tau = 0$ for reproducibility. We adopt accuracy as the metric for all tasks.

\paragraph{Data} In default settings, we set $N = 10$, $p = 0.3$, $T = 5$, and ER model for generating dynamic graphs. For each task and setting, we randomly generate one hundred problem instances for evaluation. 

We run the experiments three times with different seeds, and report the average performance and their standard deviations.

\begin{table*}[htbp]
\caption{The overall model performance (ACC\%) on the dynamic graph tasks. In the data column, `N' denotes the number of nodes in the dynamic graph. In the model column, `Random' denotes the random baseline which uniformly outputs one of the possible solutions, and `$\Delta$' denotes the performance improvement of GPT-3.5 over the random baseline.}
\label{tab:main}
\begin{tabular}{@{}ccccccccccc@{}}
\toprule
Task                    &          & \multicolumn{3}{c}{Temporal}                                                                                                                                           & \multicolumn{3}{c}{Spatial}                                                                                                                                                          & \multicolumn{3}{c}{Spatial-Temporal}                                                                                                                               \\ \midrule
Data                    & model    & \begin{tabular}[c]{@{}c@{}}when\\ link\end{tabular} & \begin{tabular}[c]{@{}c@{}}when\\ connect\end{tabular} & \begin{tabular}[c]{@{}c@{}}when\\ tclosure\end{tabular} & \begin{tabular}[c]{@{}c@{}}neighbor\\ at time\end{tabular} & \begin{tabular}[c]{@{}c@{}}neighbor\\ in periods\end{tabular} & \begin{tabular}[c]{@{}c@{}}check\\ tclosure\end{tabular} & \begin{tabular}[c]{@{}c@{}}check\\ tpath\end{tabular} & \begin{tabular}[c]{@{}c@{}}find\\ tpath\end{tabular} & \begin{tabular}[c]{@{}c@{}}sort\\ edge\end{tabular} \\ \midrule
\multirow{3}{*}{N = 5}  & GPT-3.5   & \ms{68.0}{2.8}                                      & \ms{97.7}{0.9}                                         & \ms{52.7}{2.4}                                          & \ms{86.0}{2.2}                                             & \ms{42.3}{1.7}                                               & \ms{69.0}{2.2}                                           & \ms{58.7}{2.1}                                        & \ms{79.0}{4.1}                                       & \ms{78.0}{1.4}                                      \\
                        & Random   & 3.2                                                 & 20.0                                                   & 20.0                                                    & 3.2                                                        & 3.2                                                          & 50.0                                                     & 50.0                                                  & 9.3                                                  & 13.1                                                \\
                        & $\Delta$ & +64.8                                               & +77.7                                                  & +32.7                                                   & +82.8                                                      & +39.1                                                        & +19.0                                                    & +8.7                                                  & +69.7                                                & +64.9                                               \\ \midrule
\multirow{3}{*}{N = 10} & GPT-3.5   & \ms{33.7}{2.1}                                      & \ms{77.0}{2.9}                                         & \ms{73.0}{1.6}                                          & \ms{34.0}{1.4}                                             & \ms{15.7}{4.2}                                               & \ms{66.7}{4.5}                                           & \ms{63.7}{2.6}                                        & \ms{78.3}{6.0}                                       & \ms{29.3}{4.0}                                      \\
                        & Random   & 3.2                                                 & 20.0                                                   & 20.0                                                    & 0.1                                                        & 0.1                                                          & 50.0                                                     & 50.0                                                  & 6.7                                                  & 0.0                                                 \\
                        & $\Delta$ & +30.4                                               & +57.0                                                  & +53.0                                                   & +33.9                                                      & +15.6                                                        & +16.7                                                    & +13.7                                                 & +71.6                                                & +29.3                                               \\ \midrule
\multirow{3}{*}{N = 20} & GPT-3.5   & \ms{40.3}{1.7}                                      & \ms{17.7}{4.2}                                         & \ms{63.3}{0.9}                                          & \ms{17.7}{1.7}                                             & \ms{2.0}{0.8}                                                & \ms{64.3}{7.3}                                           & \ms{57.0}{2.2}                                        & \ms{85.0}{0.8}                                       & \ms{0.0}{0.0}                                       \\
                        & Random   & 3.2                                                 & 20.0                                                   & 20.0                                                    & 0.0                                                        & 0.0                                                          & 50.0                                                     & 50.0                                                  & 7.3                                                  & 0.0                                                 \\
                        & $\Delta$ & +37.1                                               & -2.3                                                   & +43.3                                                   & +17.7                                                      & +2.0                                                         & +14.3                                                    & +7.0                                                  & +77.7                                                & 0.0                                                \\ \midrule
\multirow{3}{*}{Avg.}   & GPT-3.5   & \ms{47.3}{1.2}                                      & \ms{64.1}{0.3}                                         & \ms{63.0}{1.0}                                          & \ms{45.9}{3.1}                                             & \ms{20.0}{0.8}                                               & \ms{66.7}{2.9}                                           & \ms{59.8}{0.8}                                        & \ms{80.8}{0.3}                                       & \ms{35.8}{2.0}                                      \\
                        & Random   & 3.2                                                 & 20.0                                                   & 20.0                                                    & 1.1                                                        & 1.1                                                          & 50.0                                                     & 50.0                                                  & 7.8                                                  & 4.4                                                 \\
                        & $\Delta$ & +44.1                                               & +44.1                                                  & +43.0                                                   & +44.8                                                      & +18.9                                                        & +16.7                                                    & +9.8                                                  & +73.0                                                & +31.4                                               \\ \bottomrule
\end{tabular}
\end{table*}

\subsection{Results with data of different statistics}

We first compare GPT-3.5 on each task with different graph sizes, where $N$ is set to 5, 10 and 20 respectively. From Table~\ref{tab:main}, we have the following observations.

\textbf{\uline{Observation 1.} LLMs have preliminary spatial-temporal understanding abilities on dynamic graphs.}

As shown in Table~\ref{tab:main}, on average, GPT-3.5 has shown significant performance improvement (from +9.8\% to +73.0\%) over the baseline for all tasks, indicating that LLMs indeed understand the dynamic graph as well as the question in the task, and are able to exploit spatial-temporal information to give correct answers instead of guessing by outputting randomly generated answers. Overall, we can find that LLMs have the ability to recognize time, structures, and spatial-temporal patterns. 

\textbf{\uline{Observation 2.} Most dynamic graph tasks exhibit increasing difficulty for LLMs as the graph size grows.}

As shown in Table~\ref{tab:main}, for most tasks, the performance of GPT-3.5 drops as the graph size $N$ increases. For example, the performance drops from 97.7\% to 17.7\% on `when connect' task, and 42.3\% to 2.0\% on `neighbor in periods' task. This phenomenon may be due to two factors: 1) From the task perspective, the solution space is enlarged so that it is harder for any model to obtain the correct solution, \eg, the accuracy of the random baseline also drops significantly on `sort edge' task. 2) From the model perspective, it is harder for the model to retrieve the useful information inside the data since the input space is enlarged, \eg, on `when connect' task, the performance drops drastically while the solution space remains the same. This observation shows that it is worthy of exploring handling larger dynamic graph contexts with LLMs.

\begin{figure}
    \centering
    \includegraphics[width = 0.45\textwidth]{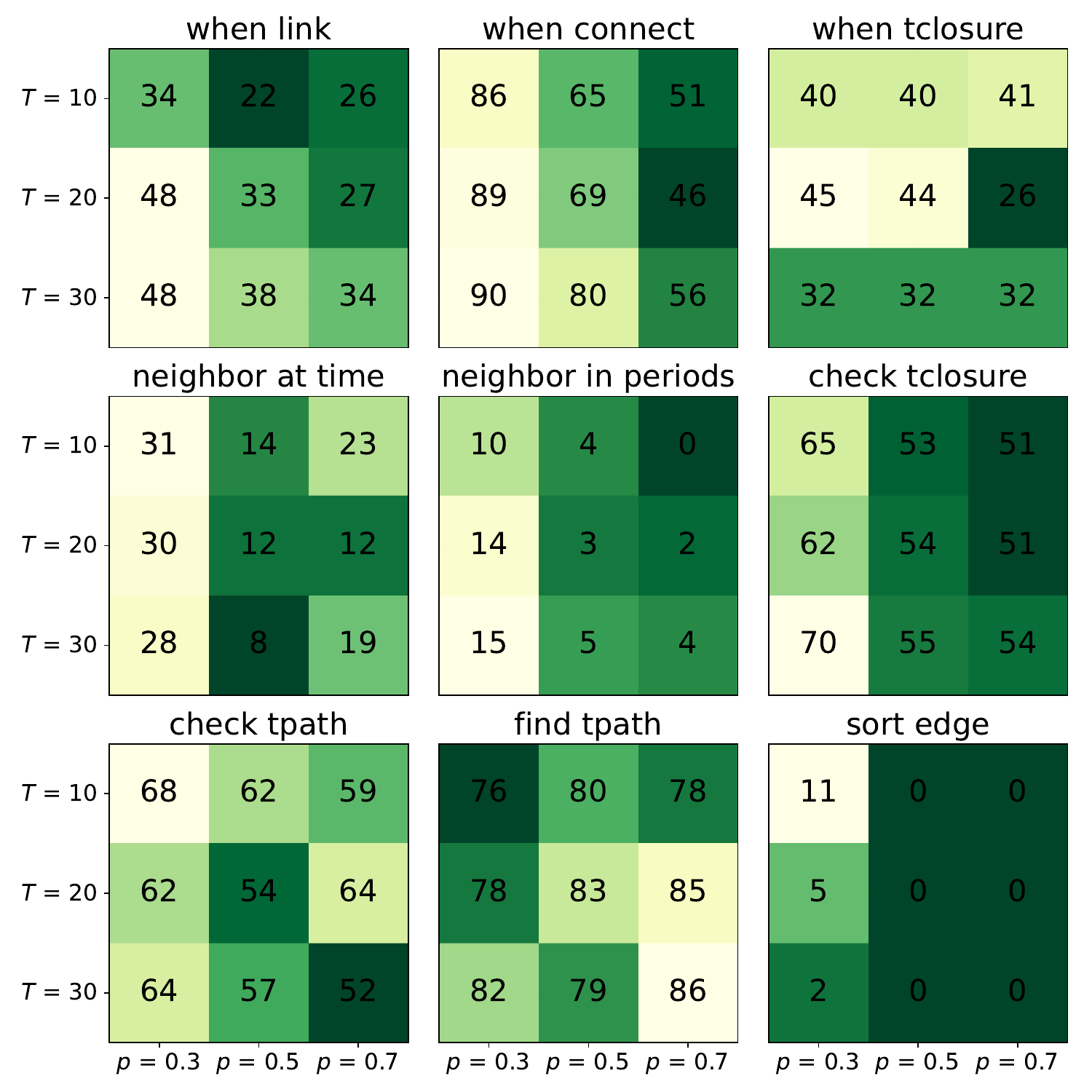}
    \caption{Performance comparisons (ACC\%) on the dynamic graph tasks with different density $p$ and time span $T$. (Best viewed in color)}
    \label{fig:Tp}
\end{figure}

\begin{table*}
\caption{Performance comparisons (ACC\%) of various prompting methods on the dynamic graph tasks. `Random' denotes the random baseline which uniformly outputs one of the possible solutions. The best result for each task is in bold.}
\label{tab:prompt}
\begin{tabular}{@{}cccccccccc@{}}
\toprule
Task                                                    & \multicolumn{3}{c}{Temporal}                                                                                                                                           & \multicolumn{3}{c}{Spatial}                                                                                                                                                          & \multicolumn{3}{c}{Spatial-Temporal}                                                                                                                               \\ \midrule
\begin{tabular}[c]{@{}c@{}}Prompt\\ Method\end{tabular} & \begin{tabular}[c]{@{}c@{}}when\\ link\end{tabular} & \begin{tabular}[c]{@{}c@{}}when\\ connect\end{tabular} & \begin{tabular}[c]{@{}c@{}}when\\ tclosure\end{tabular} & \begin{tabular}[c]{@{}c@{}}neighbor\\ at time\end{tabular} & \begin{tabular}[c]{@{}c@{}}neighbor\\ in periods\end{tabular} & \begin{tabular}[c]{@{}c@{}}check\\ tclosure\end{tabular} & \begin{tabular}[c]{@{}c@{}}check\\ tpath\end{tabular} & \begin{tabular}[c]{@{}c@{}}find\\ tpath\end{tabular} & \begin{tabular}[c]{@{}c@{}}sort\\ edge\end{tabular} \\ \midrule
zero-shot                                               & \ms{2.3}{0.5}                                       & \ms{73.3}{2.1}                                         & \ms{68.0}{0.8}                                          & \msone{36.0}{4.3}                                          & \ms{4.3}{2.1}                                                & \msone{70.7}{1.7}                                        & \msone{66.0}{5.4}                                     & \ms{56.3}{9.0}                                       & \ms{33.7}{7.4}                                      \\
one-shot                                                & \msone{33.7}{2.1}                                   & \msone{77.0}{2.9}                                      & \ms{73.0}{1.6}                                          & \ms{34.0}{1.4}                                             & \msone{15.7}{4.2}                                            & \ms{66.7}{4.5}                                           & \ms{63.7}{2.6}                                        & \ms{78.3}{6.0}                                       & \ms{29.3}{4.0}                                      \\
zero-shot COT                                           & \ms{1.0}{0.8}                                       & \ms{58.3}{1.2}                                         & \ms{70.0}{1.6}                                          & \ms{32.0}{0.8}                                             & \ms{4.3}{2.6}                                                & \ms{55.0}{1.4}                                           & \ms{62.3}{2.9}                                        & \ms{58.0}{9.1}                                       & \msone{44.7}{0.5}                                   \\
one-shot COT                                            & \ms{10.3}{0.5}                                      & \ms{76.0}{2.4}                                         & \msone{80.0}{1.6}                                       & \ms{27.7}{1.9}                                             & \ms{13.0}{3.6}                                               & \ms{57.7}{2.1}                                           & \ms{57.7}{3.4}                                        & \msone{81.3}{2.6}                                    & \ms{24.7}{2.4}                                      \\ \bottomrule
\end{tabular}
\end{table*}

We then compare GPT-3.5 on each task with different time span $T$ and density $p$, where $T$ is set to 10, 20, and 30 respectively, and $p$ is set to 0.3, 0.5, and 0.7 respectively. From Figure~\ref{fig:Tp}, we have the following observations.

\textbf{\uline{Observation 3.} For LLMs, the difficulties of dynamic graph tasks are not sensitive to the time span but sensitive to the graph density.}

As shown in Figure~\ref{fig:Tp}, for most tasks, the model performance is close as the time span $T$ increases while the density $p$ remains the same. If we keep the time span $T$ the same and increase the density $p$, the model performance drops for most tasks. One exception is the task `find tpath' where the model performance increases as the two factors increase. Another interesting finding from the heatmap is that LLMs are relatively more sensitive with the time span $T$ in temporal tasks while the density $p$ in spatial tasks, possibly due to the different points of focus for these tasks. It can be also observed in spatial-temporal tasks, where the model performance mainly changes along with the diagonal of the time span $T$ and density $p$.    

To investigate how the performance of LLMs varies when the task requires additional temporal information other than only structural information, we make comparisons with different time span $T$ and graph size $N$ on the `neighbor at time' task. We have the following observation.

\textbf{\uline{Observation 4.} Temporal information adds additional difficulties to LLMs in comparisons with static graphs.}

\begin{figure}
    \centering
    \includegraphics[width = 0.45\textwidth]{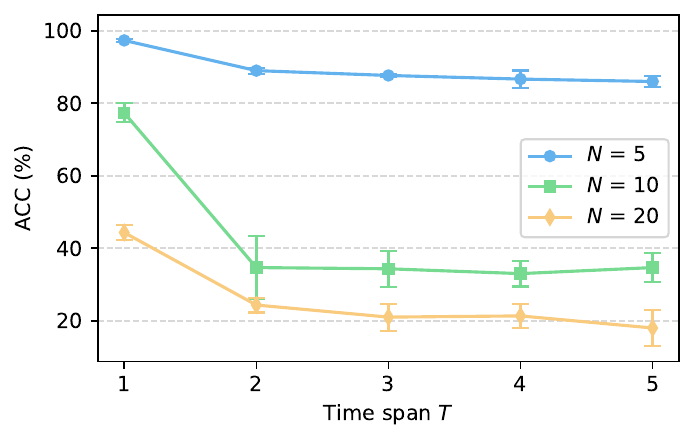}
    \caption{Performance of GPT-3.5 on the `neighbor at time' task as the time span $T$ increases with different network sizes $N$. Note that when $T=1$, the data degenerates to a static graph, since there is only one timestamp on the graph. }
    \label{fig:time}
\end{figure}

As shown in Figure~\ref{fig:time}, GPT-3.5 has a drastic performance drop when the time span $T$ increases from 1 to 2. The possible reason is that the task is changed from static to dynamic, serving as a more challenging setting, since the model has to capture the additional temporal information.  Similar to the results from Figure~\ref{fig:Tp}, the model performance is not sensitive to the time span when the task is already a dynamic graph problem. 

\subsection{Results with different prompting methods}
We then make comparisons with different prompting methods, including zero-shot prompting, one-shot prompting, zero-shot chain-of-thoughts, and one-shot chain-of-thoughts. From Tab.~\ref{tab:prompt}, we have the following observations.

\textbf{\uline{Observation 5.} General advanced prompting techniques do not guarantee a performance boost in tackling spatial-temporal information.}

As shown in Table~\ref{tab:prompt}, some advanced prompting methods like zero-shot COT and one-shot COT achieve higher performance than other prompting methods in the tasks `when tclosure', `find tpath' and `sort edge'. Note that these tasks involve more complex dynamic graph concepts or have to tackle a large time span, which shows that the chain-of-thoughts method can, to some extent, activate the model's reasoning ability by thinking step by step on complex tasks. However, no prompting methods consistently achieve the best performance on all tasks, which calls for the need to design special advanced prompting methods to boost LLMs' performance in handling spatial-temporal information on dynamic graphs.

\subsection{Results with different LLMs}

\begin{figure}
    \centering
    \includegraphics[width = 0.48 \textwidth]{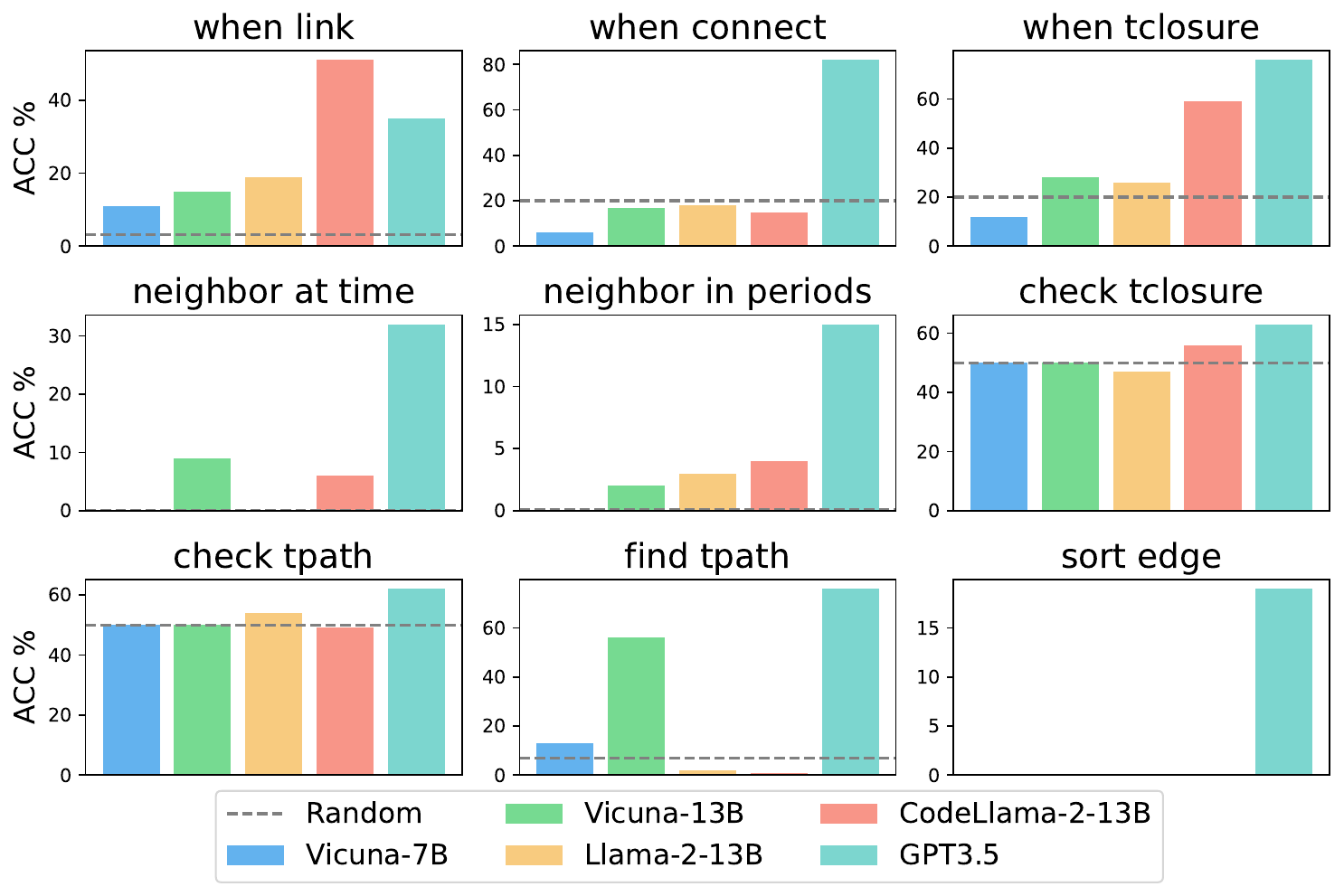}
    \caption{Performance comparisons (ACC\%) of various LLMs on the dynamic graph tasks. `Random' denotes the random baseline which uniformly outputs one of the possible solutions. (Best viewed in color)}
    \label{fig:model}
\end{figure}

\begin{table}[]
\caption{Valid rate (\%) of different LLMs. An answer is judged as valid if it meets the requirement of the answer template and can be parsed by the evaluator program.}
\label{tab:invalid}
\adjustbox{width = 0.48 \textwidth}{
\begin{tabular}{@{}ccccccc@{}}
\toprule
Valid Rate      & \begin{tabular}[c]{@{}c@{}}when\\ link\end{tabular} & \begin{tabular}[c]{@{}c@{}}when\\ connect\end{tabular} & \begin{tabular}[c]{@{}c@{}}when\\ tclosure\end{tabular} & \begin{tabular}[c]{@{}c@{}}neighbor\\ at time\end{tabular} & \begin{tabular}[c]{@{}c@{}}neighbor\\ in periods\end{tabular} & \begin{tabular}[c]{@{}c@{}}sort\\ edge\end{tabular} \\ \midrule
Vicuna-7B       & 100                                                 & 100                                                    & 100                                                     & 100                                                 & 99                                                       & 10                                                  \\
Vicuna-13B      & 92                                                  & 89                                                     & 100                                                     & 93                                                  & 97                                                       & 92                                                  \\
Llama-2-13B     & 99                                                  & 97                                                     & 95                                                      & 21                                                  & 91                                                       & 74                                                  \\
CodeLlama-2-13B & 100                                                 & 100                                                    & 100                                                     & 100                                                 & 96                                                       & 100                                                 \\
GPT-3.5          & 100                                                 & 100                                                    & 100                                                     & 100                                                 & 100                                                      & 100                                                 \\ \bottomrule
\end{tabular}
}
\end{table}

\begin{table}[]
\caption{Performance comparisons (ACC\%) of various prompting methods and dynamic graph generation models on the `when link' task. }
\label{tab:data}
\begin{tabular}{@{}cccc@{}}
\toprule
Generation Model & ER Model          & SB Model         & FF Model          \\ \midrule
zero-shot        & \ms{2.3}{0.5}     & \ms{7.7}{1.7}     & \ms{5.3}{2.5}     \\
one-shot         & \msone{33.7}{2.1} & \msone{46.0}{2.9} & \msone{48.0}{7.1} \\
zero-shot COT    & \ms{1.0}{0.8}     & \ms{5.7}{3.1}     & \ms{2.0}{1.6}     \\
one-shot COT     & \ms{10.3}{0.5}    & \ms{15.3}{0.9}    & \ms{13.0}{2.9}    \\ \bottomrule
\end{tabular}
\end{table}

\begin{table*}
\caption{Model performance (ACC\%) on the dynamic graph tasks with one-shot prompting method and our proposed \modelv prompting methods (v1 to v4). The best and the second-best results for each task are in bold and underlined respectively.}
\label{tab:adv_prompt}
\begin{tabular}{@{}cccccccccc@{}}
\toprule
Task                       & \multicolumn{3}{c}{Temporal}                                                                                                                                           & \multicolumn{3}{c}{Spatial}                                                                                                                                               & \multicolumn{3}{c}{Spatial-Temporal}                                                                                                                               \\ \midrule
Prompting methods                 & \begin{tabular}[c]{@{}c@{}}when\\ link\end{tabular} & \begin{tabular}[c]{@{}c@{}}when\\ connect\end{tabular} & \begin{tabular}[c]{@{}c@{}}when\\ tclosure\end{tabular} & \begin{tabular}[c]{@{}c@{}}neighbor\\ at time\end{tabular} & \begin{tabular}[c]{@{}c@{}}neighbor\\ in periods\end{tabular} & \begin{tabular}[c]{@{}c@{}}check\\ tclosure\end{tabular} & \begin{tabular}[c]{@{}c@{}}check\\ tpath\end{tabular} & \begin{tabular}[c]{@{}c@{}}find\\ tpath\end{tabular} & \begin{tabular}[c]{@{}c@{}}sort\\ edge\end{tabular} \\ \midrule
one-shot prompt                       & \ms{33.7}{2.1}                                      & \mstwo{77.0}{2.9}                                      & \ms{73.0}{1.6}                                          & \ms{34.0}{1.4}                                      & \ms{15.7}{4.2}                                           & \ms{66.7}{4.5}                                           & \msone{63.7}{2.6}                                     & \ms{78.3}{6.0}                                       & \ms{29.3}{4.0}                                      \\
v1: Think (about) nodes and then time & \ms{40.0}{1.6}                                      & \ms{77.0}{4.1}                                         & \msone{74.0}{1.4}                                       & \ms{34.0}{0.8}                                      & \ms{15.0}{4.2}                                           & \msone{69.3}{1.7}                                        & \ms{61.0}{3.3}                                        & \msone{79.0}{7.5}                                    & \ms{30.0}{3.6}                                      \\
v2: Think (about) time and then nodes & \ms{37.3}{2.6}                                      & \ms{76.7}{3.4}                                         & \mstwo{73.3}{0.5}                                       & \ms{31.7}{1.9}                                      & \mstwo{15.7}{3.4}                                        & \mstwo{67.0}{2.9}                                        & \ms{61.3}{1.9}                                        & \msone{79.0}{7.5}                                    & \msone{30.7}{3.9}                                   \\
v3: Pick nodes and then time  & \mstwo{59.3}{2.1}                                   & \msone{77.0}{2.4}                                      & \ms{68.0}{0.8}                                          & \mstwo{35.0}{2.9}                                   & \msone{16.7}{4.7}                                        & \ms{65.0}{3.7}                                           & \ms{62.3}{2.9}                                        & \ms{78.0}{5.4}                                       & \mstwo{30.0}{2.9}                                   \\
v4: Pick time and then nodes  & \msone{76.7}{1.7}                                   & \ms{76.3}{3.9}                                         & \ms{68.7}{0.9}                                          & \msone{35.7}{2.5}                                   & \ms{15.3}{3.3}                                           & \ms{65.3}{2.9}                                           & \mstwo{63.3}{2.6}                                     & \ms{78.3}{5.8}                                       & \ms{29.3}{2.9}                                      \\ \bottomrule
\end{tabular}
\end{table*}

We then make comparisons with different LLMs, including GPT-3.5, Llama-2-13B, Vicuna-7B, Vicuna-13B and CodeLlama-2-13B. From Figure~\ref{fig:model}, we have the following observations.

\textbf{\uline{Observation 6.} LLMs' abilities on dynamic graph tasks are related to the model scale. }

As shown in Figure~\ref{fig:model}, smaller LLMs like vicuna-7B and Llama-2-13B have performance lower than GPT-3.5 for all tasks, and even lower than random baseline for several tasks like `when connect' and `check tclosure'. Overall, for these tasks, larger LLMs have better performance. 

To further investigate whether the lower performance stems from the incompetence of understanding instructions or performing reasoning, we show the valid rate of the answers given by different LLMs in several tasks. An answer is judged as valid if it meets the requirement of the answer template and can be parsed by the evaluator program. From Table~\ref{tab:invalid}, we find that 1) smaller models have significantly lower valid rates for some tasks, \eg, 21\% of Llama-2-13B in the task `neighbor at time', demonstrating their limitations in understanding human instructions for dynamic graph tasks. 2) In some tasks, the smaller models have high valid rates, while having significantly lower performance than GPT-3.5, showing their limitations in reasoning for dynamic graph tasks. 

\textbf{\uline{Observation 7.} Training on codes may help LLMs tackle on dynamic graph tasks. }

As shown in Figure~\ref{fig:model}, compared with Llama-2-13B, CodeLlama-2-13B shows significantly better results in most tasks. In particular, CodeLlama-2-13B even outperforms GPT-3.5 in the task `when link'. Note that in comparison with Llama-2-13B, CodeLlama-2-13B is further pretrained on a large corpus of code data, which shows the potential of improving the performance of LLMs on dynamic graph tasks by training with codes. One possible reason is that the code data covers more implicit knowledge of structures and sequences,\eg, the control flows of the programs and their comments as explanations, which might be useful for LLMs to understand dynamic graphs. 

\subsection{Results with different data generators}

We make comparisons with various prompting methods and dynamic graph generation models, including Erdős–Rényi (ER) model, Stochastic Block (SB) model, and Forest Fire (FF) model, on the `when link' task. To keep the number of edges similar, we set the class number as 2, the in-class probability as 0.4, the cross-class probability as 0.2 for SB model, and the forward burning probability as 0.5 for FF model. 

\textbf{\uline{Observation 8.} General prompting methods have consistent performance with different dynamic graph generators in the same task.}

As shown in Table~\ref{tab:data}, the `one-shot' prompt method consistently achieves the best performance with different dynamic graph generation models in the  `when link' task. The results indicate that the evaluation of different prompting methods on dynamic graphs may not be closely related to the dynamic graph generators.  
\subsection{Exploring advanced dynamic graph prompts}

In this section, we aim to explore advanced dynamic graph prompting techniques to improve the reasoning ability of LLMs on dynamic graphs. The chain-of-thoughts prompting is shown as a general advanced prompting technique to activate LLMs'  complex reasoning abilities, while it does not effectively improve performance on dynamic graphs as shown in Table~\ref{tab:prompt}. 

To have further developments, we draw inspiration from dynamic graph learning literature where most works tackle spatial-temporal information separately, \eg, to tackle time first and then structures, or to tackle structures first and then time. Intuitively, this thought breaks down the complex spatial-temporal information into two separate dimensions so that the difficulty can be decreased. To this end, we propose Disentangled Spatial-Temporal Thoughts (\modelvnosp) to improve LLMs' reasoning abilities on dynamic graphs, that is to instruct LLMs to sequentially think about the nodes or time. Specifically, we design several prompts and add the prompts after the task instruction in the one-shot prompt, which are denoted as `v1' to `v4' respectively in Table~\ref{tab:adv_prompt}.

\textbf{\uline{Observation 9.} The prompting of instructing LLMs to separately tackle spatial and temporal information significantly improves the performance.}

As shown in Table~\ref{tab:adv_prompt}, the prompt `v4' achieves the accuracy of 76.7\% in the `when link' task, significantly surpassing the one-shot prompt (33.7\%), showing that guiding the LLM to handle time before nodes may help the model improve the spatio-temporal understanding ability on dynamic graphs. For spatial tasks, it seems that it would be better for the LLM to think about spatial information before temporal information (\eg, the prompt `v1' achieves 69.3\% in the `check tclosure' task). While our proposed methods provide performance gains in most tasks, there exist some tasks that are not positively affected. Designing specific prompting methods for LLMs on dynamic graphs is still an open research question.   

\section{Conclusion}
In this paper, we propose a novel \model benchmark to evaluate LLMs' spatial-temporal understanding capabilities on dynamic graphs, which remains unexplored in literature. The proposed benchmark encompasses nine specially devised tasks, which assess the capabilities of LLMs to handle both temporal and spatial information on dynamic graphs. The evaluation procedure involves a diverse range of LLMs, prompting techniques, data generators, and data statistics. We also propose Disentanlged Spatio-Temporal Thoughts (\modelvnosp) as an advanced prompting method to enhance reasoning capabilities by guiding LLMs to think about time and structures separately. Through comprehensive experiments, we provide nine fine-grained observations that would be helpful for understanding LLMs' reasoning abilities on dynamic graphs.
We hope that future work can be developed based on our proposed benchmark and observations.

\begin{acks}
This work is supported by the National Key Research and Development Program of China No. 2023YFF1205001, National Natural Science Foundation of China (No. 62222209, 62250008, 62102222, 62206149), Beijing Key Lab of Networked Multimedia and Beijing National Research Center for Information Science and Technology under Grant No. BNR2023RC01003, BNR2023TD03006.
\end{acks}

{
\small
\bibliography{ref}
\bibliographystyle{ACM-Reference-Format}
}
\appendix

\section{Additional Experiments}
\paragraph{Results of spatial tasks on real-world datasets}  We add experiments on three larger-scale real-world datasets. The details are as follows:
\begin{itemize}[leftmargin=0.5cm]
    \item Enron~\cite{shetty2004enron}, an email correspondence dataset containing emails exchanged among employees of the ENRON energy company with 50K edges.
\item DBLP~\cite{Tang:08KDD}, an academic networks dataset containing citations between papers with 100K edges. Each node is the paper published at year $y$, and each edge denotes the citation between the two paper.
\item Flights~\cite{schafer2014bringing}, a dynamic flight network illustrating the development of the air traffic during the COVID-19 pandemic with 2M edges. This dataset encompasses all flights documented by the 2500 members of the OpenSky network since January 1st, 2020. The nodes represent airports and there is an edge between two nodes at time $t$, if on day $t$ there is a flight between two airports.
\end{itemize}

We compare the performance (ACC) of static baseline methods NLGraph~\cite{wang2023can}, GPT4Graph~\cite{guo2023gpt4graph} in the spatial task 'neighbors at time'. As shown in Table~\ref{tab:real_syn}, our method has a significant performance over the baselines on larger-scale real-world dynamic graphs, which can be credited to our consideration of temporal information on dynamic graphs. Though we focus on evaluating LLMs' spatial-temporal understanding abilities on dynamic graphs, the proposed method out-performs the recent baselines for static graphs in real-world datasets.

\begin{table}[]
\caption{Task instructions in DyG prompts.}
\label{tab:task_prompt}
\begin{tabular}{p{1.3cm} | p{6.45cm}}
\toprule
Task                & Task Instruction                                                                                                                                                                                                                                                    \\ \midrule
when link           & Your task is to answer when two nodes are linked in the dynamic graph.                                                                                                                                                                                              \\
when \ \ connect        & Your task is to answer when two nodes are first connected in the dynamic graph. Two nodes are connected if there exists a path between them.                                                                                                                        \\
when \ \ tclosure       & Your task is to answer when the three nodes in the dynamic graph first close the triad. Two nodes with a common neighbor is said to have a triadic closure, if they are linked since some time so that the three nodes have linked with each other to form a triad. \\
neighbor at time    & Your task is to answer what nodes are linked with a given node at a given time in the dynamic graph.                                                                                                                                                                \\
neighbor in periods & Your task is to answer what nodes are linked with one node only after some time in the dynamic graph                                                                                                                                                                \\
check tclosure      & Your task is to answer whether three nodes in the dynamic graph formed a closed triad. A closed triad is composed of three nodes which have linked with each other some time.                                                                                       \\
check tpath         & Your task is to answer whether a path is chronological in the dynamic graph. The time of the edges in a chronological path from source node to target node must not decrease, e.g., [2, 3, 5] is a chronological path in the dynamic graph [(2, 3, 0), (3, 5, 1)]   \\
find tpath          & Your task is to find a chronological path in the dynamic graph. The time of the edges in a chronological path from source node to target node must not decrease, e.g., [2, 3, 5] is a chronological path in the dynamic graph [(2, 3, 0), (3, 5, 1)]                \\
sort edge           & Your task is to sort the edges in the dynamic graph by time from earlest to latest.                                                                                                                                                                                 \\ \bottomrule
\end{tabular}
\end{table}

\begin{table}[]
\caption{Results (ACC) on real-world datasets in the spatial task `neighbors at time'.}
\label{tab:real_syn}
\begin{tabular}{@{}llll@{}}
\toprule
Datasets  & Enron & DBLP & Flights \\ \midrule
NLGraph   & 0.19   & 0.33  & 0.42     \\
GPT4Graph & 0.30   & 0.35  & 0.47     \\
Ours      & 0.45   & 0.43  & 0.47     \\ \bottomrule
\end{tabular}
\end{table}

\paragraph{Comparisons of static baselines on spatial tasks} We compare the static baselines~\cite{wang2023can,guo2023gpt4graph} on spatial tasks. As shown in Table~\ref{tab:syn_base}, our method out-performs the recent static baselines, showing the effectiveness of our method for tackling problems on dynamic graphs.

\begin{table}[]
\caption{Comparisons (ACC) on several spatial tasks.}
\label{tab:syn_base}
\begin{tabular}{@{}lcc@{}}
\toprule
Task      & neighbor at time & neighbor in periods \\ \midrule
NLGraph   & 0.18              & 0.06                  \\
GPT4Graph & 0.22              & 0.12                 \\
Ours      & 0.34              & 0.16                 \\ \bottomrule
\end{tabular}
\end{table}

\begin{table}[]
\caption{Comparisons on link prediction tasks.}
\label{tab:link}
\begin{tabular}{@{}lllll@{}}
\toprule
Dataset   & \multicolumn{2}{l}{Enron} & \multicolumn{2}{l}{DBLP} \\ \cmidrule(l){2-3}  \cmidrule(l){4-5}
Metric    & F1         & Recall       & F1         & Recall      \\ \midrule
NLGraph   & 0.11       & 0.12         & 0.26       & 0.29        \\
GPT4Graph & 0.17       & 0.22         & 0.27       & 0.35        \\
Ours      & 0.29       & 0.60         & 0.33       & 0.74        \\ \bottomrule
\end{tabular}
\end{table}

\paragraph{Links prediction tasks on real-world datasets} we add experiments for link prediction tasks to further compare the baselines~\cite{wang2023can,guo2023gpt4graph} for predictive tasks. As shown in Table~\ref{tab:link}, our method out-performs static baselines NLGraph and GPT4Graph with a large margin, demonstrating the effectiveness of handling spatial-temporal information on dynamic graphs. 

\begin{table}[]
\caption{Comparisons (ACC) with different time formatting types on several tasks.}
\label{tab:time}
\begin{tabular}{@{}ccc@{}}
\toprule
Time types & when tclosure & check tclosure \\ \midrule
Original   & 0.59            & 0.56             \\
UNIX       & 0.35            & 0.53             \\
Date       & 0.47            & 0.57             \\ \bottomrule
\end{tabular}
\end{table}

\paragraph{Effects of time formatting types.} We add experiments with different types of timestamps, including original, UNIX timestamp, Date, using Codellama2-13B and two tasks. The results are shown in the table~\ref{tab:time}. The original timestamp is an integer ranging from 1 to $T$, where T is the time span. The UNIX timestamp ranges from 2010-01-01 00:00:00 to 2020-01-01 00:00:00. The date is formatted as `\%Year\%Month\%Day', \eg, `20200101'. The results show that using UNIX timestamp or date for time formatting reduces the performance, which may be due to the increased complexity for LLMs to infer the ordering between time.

\begin{table}[]
\caption{Comparisons (ACC) with different node formatting types on several tasks.}
\label{tab:node}
\begin{tabular}{@{}ccc@{}}
\toprule
Node name types & when tclosure & check tclosure \\ \midrule
Original        & 0.59            & 0.56             \\
Random indexes  & 0.62            & 0.55             \\
People names    & 0.68            & 0.62             \\ \bottomrule
\end{tabular}
\vspace{-10pt}
\end{table}

\paragraph{Effects of node formatting types.} We add experiments with different types of node names, including original, random indexes, people names, using Codellama2-13B and two tasks. The results are shown in the table~\ref{tab:node}. The original node formatting uses integers ranging from 0 to $N$ to represent the nodes, where $N$ refers to the number of nodes, while the random indexes adopt integers ranging from 0 to 1e8. For the formatting of people names, several names are adopted to represent nodes, \eg, 'Aiden', 'Priya', 'Dmitri'. One interesting phenomenon is that using names to represent nodes may improve the performance, which may be due to that LLMs may be more familiar with names than integers. The results are consistent with previous literature\cite{fatemi2023talk}.

\section{Implementation Details}
\paragraph{Task prompts} For each problem instance, the prompt is constructed by a unified template including DyG instructions describing a dynamic graph, task instructions describing the task, answer instructions describing the answer template, examplars (in few-shot learning) and questions. An example is illustrated in Table~\ref{tab:example} and we provide the task instruction for each task in Table~\ref{tab:task_prompt}.


\section{Additional Related Works}

\paragraph{Disentangled Representation Learning}
Disentangled representation learning seeks to identify and clarify the distinct latent factors underlying observable data, with each factor represented as a unique vector~\cite{bengio2013representation, wang2023disentangled}. These factors are crucial for unraveling the intrinsic processes shaping data formation and for generating robust representations for subsequent applications. This approach has demonstrated its utility in various fields, including computer vision~\cite{hsieh2018learning, ma2018disentangled, chen2016infogan, wang2023mixup, chen2023disenbooth, chen2024disenstudio}, and graph representation learning~\cite{ma2019disentangled, liu2020independence, yang2020factorizable, chen2021curriculum, li2021intention, li2021disentangled, li2022disentangled, zhang2023unsupervised,zhang2024disentangled,li2024disentangled,wang2023curriculum,wang2022disentangled}. ~\citet{pan2023context}  explores and evaluate the in-context learning in large language models by disentangling the effects of task recognition and task learning. \citet{qin2023disentangled} integrates tailored disentangled graph neural network layers to capture complex structural relationships within text for better performance and interpretability.
In this paper, we focus on studying the spatial-temporal understanding abilities of LLMs on dynamic graphs, and explore a disentangled prompt to improve performance via letting the model think of spatial and temporal dimensions separately. 

\paragraph{Chain-of-Thoughts} The concept of chain-of-thoughts (CoT)~\cite{wei2022chain} has garnered significant attention in recent years. CoT refers to the process by which a model generates intermediate reasoning steps that lead to the final answer, thereby mimicking human-like reasoning patterns, which has been shown to improve the performance of language models on complex tasks that require multi-step reasoning. Subsequent research has built upon these findings, exploring various techniques to optimize CoT prompting and extend its applicability to a broader range of tasks ~\cite{kojima2022large,wang2022self,zhou2023teaching,zhang2023multimodal}. ~\citet{besta2024graph} model the information generated
by an LLM as an arbitrary graph, enhancing thoughts using feedback loops. ~\citet{zhang2022automatic} design an automated process for generating CoT prompts. Since the enhanced CoT in our framework still have mixed results, one possible future direction is design automated CoT with automated graph techniques~\cite{guan2021autogl,qin2021graph,qin2022bench,zhang2022learning,li2024causalaware}. We also leave extending the framework to recently released LLMs~\cite{jiang2023mistral,bai2023qwen,du2022glm,yang2023baichuan} in future works.

\end{document}